\documentclass[11pt]{article}

\usepackage{arxiv}
\usepackage{algpseudocode}
\usepackage{algorithm}
\usepackage{graphicx}
\usepackage[english]{babel}
\usepackage{amsmath}
\usepackage{amssymb}
\usepackage{subcaption} % Kept this one, removed subfig
\usepackage{adjustbox}
\usepackage[colorlinks,allcolors=blue]{hyperref}

\usepackage[T1]{fontenc}    % use 8-bit T1 fonts
\usepackage{url}            % simple URL typesetting
\usepackage{booktabs}       % professional-quality tables
\usepackage{amsfonts}       % blackboard math symbols
\usepackage{nicefrac}       % compact symbols for 1/2, etc.
\usepackage{microtype}      % microtypography
\usepackage{lipsum}
\usepackage{color}
\usepackage{float}
\usepackage{titlesec}
\usepackage{multirow}       % Consolidated duplicate
\usepackage{afterpage}
\usepackage{xspace}         % ADDED: Required for your \xspace commands below

\usepackage[dvipsnames]{xcolor}

\definecolor{azulito}{HTML}{5E7BE6}
\definecolor{amarillito}{HTML}{E6D25E}
\definecolor{marroncito}{HTML}{E6B15E}

\usepackage{makecell} % Para saltos de línea dentro de las celdas (si fuera necesario)
\usepackage{geometry} % Para ajustar los márgenes de la página si la tabla es muy ancha
\usepackage{lineno}

\title{MoEITS: A Green AI approach for simplifying MoE-LLMs} %% Article title

\author{
  Luis Balderas \\
  {\small Department of Computer Science}\\
  {\small and Artificial Intelligence}\\
  {\small DiCITS, iMUDS, DaSCI}\\
  {\small University of Granada, Granada, Spain 18071}\\
  \texttt{luisbalru@ugr.es} \\
\And
  Miguel Lastra \\
  {\small Department of Software Engineering}\\
  {\small DiCITS, iMUDS, DaSCI}\\
  {\small University of Granada, Granada, Spain 18071}\\
  \texttt{mlastral@ugr.es} \\ 
\And
  José M. Benítez\\
  {\small Department of Computer Science}\\
  {\small and Artificial Intelligence}\\
  {\small DiCITS, iMUDS, DaSCI}\\
  {\small University of Granada, Granada, Spain 18071}\\
  \texttt{J.M.Benitez@decsai.ugr.es} \\
}

\begin{document}

\maketitle

%% Abstract
\begin{abstract}
Large language models are transforming all areas of academia and industry, attracting the attention of researchers, professionals, and the general public. In the trek for more powerful architectures, Mixture-of-Experts, inspired by ensemble models, have emerged as one of the most effective ways to follow. However, this implies a high computational burden for both training and inference. To reduce the impact on computing and memory footprint as well as the energy consumption, simplification methods has arisen as very effective procedures.

In this paper, an original algorithm, MoEITS, for MoE-LLMs simplification is presented. The algorithm is characterized by a refined simplicity, underpinned by standardized Information Theoretic frameworks. MoEITS is analyzed in depth from theoretical and practical points of view. Its computational complexity is studied. Its performance on the accuracy of the simplified LLMs and the reduction rate achieved is assessed through a thoroughly designed experimentation. This empirical evaluation includes a comparison with state-of-the-art MoE-LLM pruning methods applied on Mixtral $8\times7$B, Qwen1.5-2.7B, and DeepSeek-V2-Lite. The extensive experimentation conducted demonstrates that MoEITS outperforms state-of-the-art techniques by generating models that are both effective across all benchmarks and computationally efficient.

The code implementing the method will be available at \url{https://github.com/luisbalru/MoEITS}.
\end{abstract}

%% Keywords

%% Add \usepackage{lineno} before \begin{document} and uncomment 
%% following line to enable line numbers
%% \linenumbers

%% main text
%%

\section{Introduction}
\label{intro}

Large language models (LLMs) have disrupted the information society, revolutionizing all sectors, from a purely research perspective, with use cases in natural language processing \cite{FARHANGIAN2024102140, LIU2024102300} or computer vision \cite{ISHMAM2024102270, XIAO2024102204}; to applied research, including medicine \cite{HE2025102963}, smart cities \cite{ ZOU2025102606, ZHANG2024102038}, or finance \cite{BAO2025102616}, among others. In particular, Mixture-of-Experts LLMs emerge, inspired by traditional ensemble models, to significantly augment the capabilities of LLMs \cite{cai2024surveymixtureexperts}.

However, the quantity of hardware resources required to train and deploy models such as MoE-LLMs is immense. Furthermore, their use constitutes a significant impact on the environment \cite{10.1007/978-3-658-45010-6_15, https://doi.org/10.1002/widm.1507, 10765824}. In parallel with the rapid growth of computationally intensive AI systems, the Green AI paradigm \cite{10.1145/3381831} has emerged, which aims to build more sustainable artificial intelligence systems through the use of metrics that consider not only accuracy but also the number of parameters, electricity consumption, or the number of model operations. Among these, simplification algorithms stand out as a fundamental tool for reducing the size of pre-trained models. In this article Mixture of Expert Information Theory Simplifier, MoEITS, is proposed, a novel simplification method for MoE-LLMs based on information theory. Specifically, normalized mutual information (NMI) is used as a metric to quantify the level of redundancy present in the experts of each network block. MoEITS is an algorithm that receives a parameter $\tau$, which indicates the intensity with which the simplification of the models is carried out.

\begin{figure}[H]
	\centering
	\includegraphics[width=\textwidth]{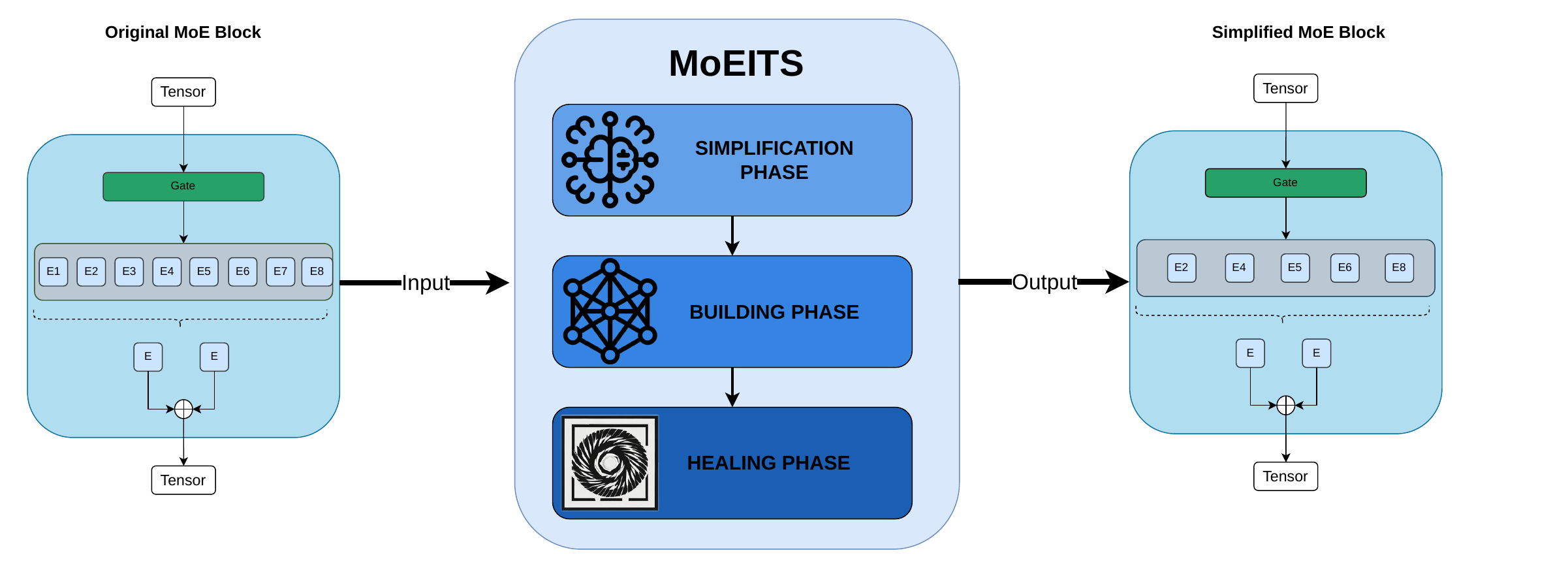}
	\caption{Diagram of MoEITS. Starting from a block of model experts, the redundancy analysis metric between experts based on normalized mutual information is applied. Next, the most relevant experts are simplified and a new version of the model is generated by simplifying, inheriting the knowledge of the original model by means of the weight matrices. Note that the layers immediately preceding and following the simplified expert block must be adjusted.}
	\label{fig:moeits}
\end{figure} 

To measure the effectiveness of MoEITS, we have designed and carried out an extensive experimentation, including the simplification of MoE models such as Mixtral $8\times7$B \cite{jiang2024mixtralexperts}, Qwen1.5-MoE-2.7B \cite{qwen_moe}, and DeepSeek-V2-Lite \cite{dai2024deepseekmoeultimateexpertspecialization}, and benchmarks to evaluate the reasoning capacity of LLMs, such as BoolQ \cite{clark-etal-2019-boolq}, HelloSwag \cite{zellers-etal-2019-hellaswag}, WinoGrande \cite{10.1145/3474381}, ARC-e and ARC-c \cite{clark2018thinksolvedquestionanswering}. The main goal of this article is to address the following research questions (RQ):

(RQ1) To what extent are information theory metrics effective in measuring expert redundancy within a MoE?

(RQ2) Can an algorithmic methodology be defined that utilizes information theory to construct simplified MoE-LLMs?

(RQ3)  At what point in the simplification process of a MoE, through the elimination of its experts, does the model cease to exhibit reasoning capabilities?

The main contributions of the article can be summarized as follows:

\begin{itemize}
	\item A methodology based on normalized mutual information is proposed to detect redundancy among experts within a block and generate a simplified MoE.
	\item Both a theoretical analysis of computational complexity and an empirical analysis based on the simplification of Mixtral $8\times7$B, Qwen1.5-2.7B and DeepSeek-V2-Lite, evaluated on the most established LLM reasoning benchmarks, are performed.
	\item An ablation study is conducted to measure the evolution of the generative capacity of Qwen1.5-2.7B applied to the BoolQ benchmark when different levels of simplification are applied.
\end{itemize}

The remainder of the article is organized as follows: Section 2 presents various state-of-the-art approaches for the simplification of MoE-LLMs, including those based on quantization and pruning. Section 3 provides a theoretical introduction to MoE-LLMs and the main current models. Section 4 presents MoEITS. Section 5 contains the empirical evaluation. Section 6 includes a discussion of the results, emphasizing the difference between simplification approaches and an ablation study. Finally, Section 7 presents the conclusions of the article.

%% Use \subsection commands to start a subsection.
\section{Related Work}
\label{related}

As with other approaches related to neural network simplification, especially for language models, quantization and pruning are the
two most prominent paradigms \cite{WOS:001616292100001}. In the case of MoE models, this
necessity becomes even more pressing, given the substantial increase in the number of parameters resulting
from the introduction of expert submodels. The following sections present the state-of-the-art
techniques in each of these two disciplines.

\subsection{Quantization techniques}
Quantization has emerged as a promising technique for compressing LLMs, particularly for deployment 
in resource-constrained environments, by reducing the precision of model weights and activations \cite{DBLP:journals/corr/abs-2404-14047, 10.5555/3666122.3666563, liao-etal-2024-apiq}. While various quantization approaches exist --ranging from model-level techniques like PolarQuant \cite{han2025polarquantquantizingkvcaches} to vector quantization methods such as TurboQuant \cite{zandieh2025turboquantonlinevectorquantization}-- this section explores and compares several recent techniques designed specifically for MoE-LLMs, grouping them according to their underlying principles.

Post-Training Quantization (PTO) methods offer a streamlined approach by quantizing
model weights after the training phase, eliminating the need for further training. This simplicity
makes them computationally inexpensive and highly attractive. Xiao \emph{et al.} introduce in \cite{xiao2023smoothquant} a method called Smooth Quant, which tackles the challenge of quantizing activations, which are often more difficult due to outliers, by redistributing the quantization
difficulty between weights and activations. It achieves this by scaling both, effectively smoothing the activations to make them more quantization-friendly. Another PTQ method, BILLM \cite{10.5555/3692070.3692876}, focuses on aggressively compressing LLM weights down to
approximately 1 bit while preserving accuracy. BILLM achieves this extreme compression through a
combination of strategies. First, it structurally selects salient weights based on Hessian
metrics and employs a residual approximation to maintain the dynamic range of these crucial
weights. Second, it utilizes an optimal splitting strategy for the remaining non-salient
weights to minimize binarization errors, leveraging the observation that these weights often
follow a bell-shaped distribution. Both Smooth Quant and BILLM share the goal of preserving
the most important information within the model during quantization, but they differ in
their specific approaches. Smooth Quant emphasizes the importance of activation scaling,
while BILLM prioritizes the identification and preservation of salient weights through
Hessian metrics and residual approximation.

Quantization-Aware Training (QAT) methods integrate the quantization process directly into the training phase, allowing the model
to adapt to the effects of quantization and potentially achieve superior accuracy compared to PTQ methods. PB-LLM \cite{yuan2024pbllm} exemplifies this approach by exploring how to optimally
identify salient weights and reintegrate them into the model through both PTQ and QAT. This hybrid
approach allows PB-LLM to achieve substantial compression while maintaining the language
reasoning capabilities of the original, larger model. PB-LLM leverages the concept
of mixed-precision quantization, assigning varying bit-widths to different parts of the model
based on their relative importance. However, the specific strategies for determining these
bit-widths and managing different model components (e.g., experts vs. tokens) distinguish
these methods.     

Mixed-precision quantization methods strategically assign different bit-widths to various parts of
the model according to their sensitivity to quantization or their overall importance. MC-MoE
\cite{huang2025mcmoe} embodies this strategy by allocating varying bit-widths to experts within
the MoE architecture based on their significance. This targeted approach allows for high
compression rates. Furthermore, MC-MoE incorporates online dynamic pruning, identifying and
pruning less important tokens and dynamically selecting experts during inference. This
combination of mixed-precision quantization and dynamic pruning further optimizes
computational efficiency, setting MC-MoE apart from other methods. Finally, Activation-Aware Weight Quantization (AWQ) \cite{MLSYS2024_42a452cb} introduces a novel approach to weight quantization by focusing on the
distribution of activations rather than solely on weight magnitudes. This method identifies and
protects the most salient weights by scaling them based on activation distributions. By focusing
on activations, AWQ avoids the need for costly retraining or reconstruction steps, thus
preserving the generalization capabilities of LLMs across different tasks and modalities. This
activation-centric perspective distinguishes AWQ from methods relying on weight magnitudes or
Hessian metrics.

\subsection{Pruning Mixture-of-Experts LLMs}
The compression of MoE LLMs via pruning techniques has emerged as a critical area of research \cite{ashkboos2024slicegpt, ma2023llmpruner}. Several novel methodologies have been proposed, each with distinct approaches to identifying and removing redundant or less impactful experts and parameters. A common thread across these studies is the goal of balancing model compression with the preservation of task performance.

One cluster of approaches focuses on expert-level pruning, with variations in how experts are deemed redundant. Lu \emph{et al.}~\cite{lu-etal-2024-experts}  explore both task-agnostic and task-specific expert pruning, demonstrating significant efficiency gains on the Mixtral $8\times7$B model. Zhang \emph{et al.}~\cite{zhang2024diversifyingexpertknowledgetaskagnostic} propose a two-stage task-agnostic method that groups and merges similar experts based on feature representations, showcasing its effectiveness across various MoE architectures. These methods underscore the importance of identifying and eliminating redundant experts to reduce computational overhead.

A more refined strategy is presented by Lee \emph{et al.}~\cite{lee-etal-2025-stun}, who introduce STUN (Structured-Then-Unstructured Pruning), a method that first applies structured pruning at the expert level, followed by unstructured pruning. This counterintuitive approach leverages behavioral similarity among experts to achieve superior compression rates and performance. Complementing these approaches, Xie \emph{et al.}~ \cite{xie2024moepruner} introduce MoE-Pruner, which identifies and removes unimportant weights based on magnitude, input activations, and router weights, further enhanced by expert-wise knowledge distillation. Another method based on expert selection is proposed by Chen \emph{et al.}~\cite{chen-etal-2025-eac}. The method combines the perspectives of quantization and pruning to avoid the expert-selection bias caused by low-bit quantization and the latency generated by non-crucial experts in certain tasks.

Furthermore, Yang \emph{et al.}~\cite{yang-etal-2024-moe} present MoE-$I^2$, a two-stage framework that combines inter-expert pruning using genetic search with intra-expert decomposition via low-rank approximation, offering a task-agnostic structured compression. Besides, Chowdhury \emph{et al.}~propose in \cite{pmlr-v235-chowdhury24a} a method which selectively removes experts that not relevant for specific tasks. Therefore, it reduces memory requirements and runtime costs. Feng \emph{et al.}~\cite{feng-etal-2025-dive} present DIVE, a diversity-enhanced reconstruction method which includes pruning and reassembly dense modules. After that, the model is retrained. Nonetheless, it is evaluated only over TinyLlama-1.1B. Finally, Gao \emph{et al.}~\cite{gao2025tomoeconvertingdenselarge} introduce ToMoE, a dynamic structural pruning technique that converts dense LLMs into MoE models by selectively activating model parameters, achieving sparse and efficient computation.

Collectively, these studies advance the field by proposing diverse pruning techniques tailored to MoE architectures, addressing the challenges of model compression while maintaining or even enhancing performance. The choice of pruning method depends on the specific requirements of the application, including the balance between compression rate, computational efficiency, and task-specific performance.

\section{Preliminaries: Mixture-of-Experts Large Language Models}
\label{preliminaries}

Mixture-of-Experts (MoE) models, first introduced in \cite{6797059}, are models containing multiple
feed-forward neural networks (FFN), acting as experts, that independently process a complete dataset.
Following this processing, the best experts for each inference are dynamically selected by means of
a router that assigns weights to the experts' contributions. Concretely, a MoE architecture 
comprises a set of independent FFNs called experts, $\{E_1, E_2, \dots, E_n\}$, and a router network or gate $G$. The output is calculated as follows:

\begin{equation}
	F(x) = \sum_{i=1}^{n} g_i E_i(x),
\end{equation} 
\noindent with
\begin{equation}
	g_i = \left\{ \begin{array}{lcc} s_i, & s_i \in \text{TopK}(\{s_i, 1 \leq i \leq n\}, K) \\
		\\
		0, & \text{otherwise} \end{array} \right.
\end{equation}
\begin{equation}
	G(x) = [s_1, s_2, \dots, s_n]
\end{equation}
\noindent where $g_i$ represents the score for the $i$-th expert, $G(x)$ is the output of the gate,
corresponding to the affinity between the input token and each expert, and $\text{TopK}(\cdot,K)$ yields the $K$ highest values among the experts' scores for each input $x$ 
(\cite{pei2025cmoefastcarvingmixtureofexperts}).

Although the MOE structure was employed for the design of Recurrent Neural Networks \cite{shazeer2017}, it has gained significant popularity in recent years due to its application in Transformer models, including both encoder-decoder and decoder-only architectures, for natural language applications. Some of the most relevant MoE architectures for LLMs are Mixtral $8\times7$B \cite{jiang2024mixtralexperts}, Google Switch-Transformers \cite{10.5555/3586589.3586709}, Qwen1.5 MoE \cite{qwen_moe}, and DeepSeek MoE \cite{dai2024deepseekmoeultimateexpertspecialization}. 

Google's Switch Transformers models address the computational challenges associated with scaling transformer-based language networks by introducing a simplified routing mechanism within the MoE framework. Rather than employing complex gating networks to determine expert selection, Switch Transformers utilize a single ``switch'' layer, which directs each token to a single expert. This simplification significantly reduces routing overhead, thereby enabling the training of extremely large MoE models with greater efficiency. The core innovation of Switch Transformers lies in their ability to maintain high model capacity while minimizing computational cost. By directing each token to a single, chosen expert, the computational burden is distributed across the available experts, allowing for the scaling of model parameters without a proportional increase in processing demands. This approach facilitates the training of models with orders of magnitude more parameters than traditional dense transformers, leading to improved performance on a variety of language tasks. Furthermore, the simplified routing mechanism contributes to increased training stability and reduced communication overhead, making Switch Transformers a practical and effective approach for scaling language models to unprecedented sizes. In essence, Google's Switch Transformers provide a pathway to efficient and scalable language modeling through the judicious application of a simplified MoE architecture.

Mixtral $8\times7$B represents a highly acclaimed and widely adopted MoE architecture, establishing itself as a de-facto benchmark within both industry and academia, particularly in
studies concerning the simplification of MoE architectures. This decoder-only model comprises
eight expert per layer, yet activates only two at any given inference step.
Consequently, the number of active parameters per token is significantly lower compared to other
Large Language Models (LLMs), such as Llama 2 70B \cite{touvron2023llama2openfoundation} (13B versus 70B, respectively), while
achieving superior performance. This efficient parameter utilization underscores Mixtral $8\times7$B's ability to deliver high-quality outputs with reduced computational demands,
facilitating its application in resource-constrained environments and its adoption as a
standard for evaluating and advancing MoE methodologies.

DeepSeek AI's DeepSeek-V2-Lite model marks a notable progression in the field of large language models, particularly through its refined implementation of the Mixture-of-Experts (MoE) architecture. The model's core innovation resides in its capacity to achieve a sophisticated balance between high-level performance and computational efficiency, a feat primarily realized through the strategic deployment of sparse activation within its network. The model's architectural design centers on a sparse MoE framework, which, unlike dense models, selectively activates specific subsets of its parameters during each processing cycle. This selective activation allows for the maintenance of a substantial overall parameter count—16 billion—while simultaneously mitigating the computational overhead typically associated with such large-scale models. DeepSeek AI has further augmented the MoE paradigm through the introduction of fine-grained expert segmentation and shared expert isolation. These methodological advancements are engineered to foster a higher degree of specialization among the model's constituent experts, thereby enhancing the granularity and fidelity of its knowledge representation.

Finally, Qwen1.5-2.7B, developed by Alibaba Cloud, distinguishes itself by its focus on balancing performance with computational efficiency, a critical consideration for deploying large models in practical applications. The architecture of Qwen1.5-2.7B leverages the MoE paradigm to selectively activate a subset of its parameters during each inference, thereby mitigating the computational cost associated with dense transformer models. This approach allows for the model to possess a substantial parameter count, enhancing its capacity to capture complex linguistic patterns, while maintaining a manageable computational footprint. Furthermore, the model incorporates advancements in routing mechanisms and expert management, contributing to improved stability and performance across diverse language tasks.
\section{Our proposal}
\label{proposal}

In this paper, we introduce MoEITS, a method for simplifying Mixture-of-Experts Large Language Models. To achieve this simplification, we employ tools from information theory \cite{1188572}, specifically
Normalized Mutual Information (NMI) \cite{doi:10.1137/0119020, e19110631}, to quantify the redundancy present within each expert block
of the network. This analysis enables the generation of a refined model where experts exhibit
complementary behavior during inference tasks, effectively reducing redundancy. Despite its
inherent power, MoEITS is remarkably easy to use, requiring only one parameter, the
threshold $\tau$, which indicates the level or intensity of simplification applied to the original model.
Furthermore, to ensure that the simplified models perform at an optimal level, MoEITS conducts the healing phase with a lightweight retraining, resulting in minimal
computational demands for its application. More details can be found in Algorithm \ref{alg:method} and Figure \ref{fig:moeits}.

\begin{algorithm}
	\caption{MoEITS method}\label{alg:method}
	\begin{algorithmic}[1]
		\Procedure{MoEITS}{model, $\tau$}
		\State Get NMI metrics from experts.
		\State Build the simplified model using NMI metrics and $\tau$. \Comment{\textsf{[Algorithm \ref{alg:simpmodel}]}}
		\State Add model's weights to the simplified model.
        \State Retrained the simplified model.
		\State \textbf{return} simplified model
		\EndProcedure
	\end{algorithmic}
\end{algorithm}

\subsection{Using Normalized Mutual Information to reduce redundancy among experts}
\label{nmi}
Information theory, introduced by Claude Shannon in \cite{1188572}, is a branch of mathematics dedicated to the study of the quantification and communication of information. One of the most important metrics in information theory is entropy, which measures the amount of uncertainty or information associated with the potential states of a random variable \cite{6773024}. If $p(x)$ is the probability of the event $x \in X$, the entropy of a discrete random variable $X$ is calculated as

\begin{equation}
	H(X) = - \sum_{x} p(x) \log p(x)
\end{equation}

In other words, $H(X)$ indicates how much information can be found on average in one vector $X$. Similarly, joint entropy, denoted as $H(X,Y)$, can be defined as the amount of uncertainty inherent in two random variables, $X$ and $Y$, considered together.

Another fundamental tool within information theory is mutual information, which is a symmetric metric that quantifies the statistical information shared between two distributions \cite{10.5555/1146355}. Mutual information is calculated as:

\begin{equation}
	I(X,Y) = H(X)+H(Y)-H(X,Y)
\end{equation}

Given that $I(X,Y)$ is not upper-bounded, in order to enhance interpretability and comparability, it would be desirable to have a normalized version of $I(X,Y)$, that is, one that takes values between 0 and 1. Following the notation proposed in \cite{nmi}, we define Normalized Mutual Information (NMI) as:

\begin{equation}
	\text{NMI}(X,Y) = \frac{I(X,Y)}{\sqrt{H(X)H(Y)}}
\end{equation}

In light of the foregoing, NMI is a highly useful metric for quantifying the amount of shared information between two distributions or vectors. Furthermore, given that it takes values between 0 and 1, it is straightforward to compare results across different vectors. Specifically, given two vectors $X$ and $Y$, a high value of $\text{NMI}(X,Y)$ implies that $X$ and $Y$ share a significant amount of information or, in other words, possess redundant information. Conversely, a low value of $\text{NMI}(X,Y)$ implies that $X$ and $Y$ share little information in common.

Considering that this work aims to simplify MoE-LLMs by identifying and eliminating redundant experts (RQ1), and that each expert is defined by a specific number of layers, whose information and learning are expressed in a weight matrix, the MoEITS method employs NMI to measure the amount of redundant information in the matrix expressions of their weights. Suppose that within a block, we have two experts $e_1$ and $e_2$, defined by respective Linear layers  $L_1^{e_1}$ and $L_2^{e_1}$ for $e_1$; $L_1^{e_2}$ and $L_2^{e_2}$ for $e_2$. Let $W_{L_i}^{e_j}, \text{with } i,j \in \{1,2\}$ the weight matrix for each Linear layer and expert. The level of redundancy $R$ between the two experts is defined as:
\begin{gather}
	r_1 = \text{NMI}(W_{L_1}^{e_1}, W_{L_1}^{e_2})\\
	r_2 = \text{NMI}(W_{L_2}^{e_1}, W_{L_2}^{e_2})\\
	R = \frac{r_1+r_2}{2} \label{eq:R}
\end{gather}

In general, for every pair of experts $e_k, e_o$ within a block, if each expert is formed by $n$ layers
\begin{equation}
	R = \frac{\sum_{i=1}^{n}r_i}{n},
\end{equation}
\noindent with
\begin{equation}
	r_i = \text{NMI}(W_{L_i}^{e_k}, W_{L_i}^{e_o})
\end{equation}

The higher the value of $R$, the more redundant the experts are. The preceding decisions assume that, as is generally the case in most models, experts share the same structure. However, if this premise were not met, these definitions could be adjusted for size, and, in the case of expression (\ref{eq:R}), introduce a weighted mean based on relevance. For simplicity, the arithmetic mean is used. 
%\fixme{(luisbalru)Introducido párrafo para incluir los detalles de las notas siguientes}
%\fxnote{Estas definiciones asumen que la estructura de los expertos es similar, al menos en número de capas lineales. No siempre tiene por qué ser así}.
%\fxnote{(luisbalru) Los modelos que he conocido hasta ahora tienen todas las capas de expertos iguales, pero sí, añado esa apreciación}

%\fxnote{La expresión de (\ref{eq:R}) es acertada por simplificada, pero podría ser más relevante una media ponderada donde cada $r_i$ tuviese un peso distinto; por ejemplo, basado en la relevancia o tamaño,}
%\fxnote{(luisbalru) De acuerdo}

\subsection{Experts pruning process}

The previous section elucidates how the MoEITS method employs NMI to quantify expert redundancy. This step corresponds to step 4 of Algorithm \ref{alg:simpmodel}. This section focuses on how experts are selected using NMI (step 5 of Algorithm \ref{alg:simpmodel} and Algorithm \ref{alg:expertsimp}).

Given a block of experts within a layer of a MoE, the application of normalized mutual information to all pairs of experts generates a symmetric matrix, which will be referred to as the NMI matrix, $\mathcal{R} = [r_{ij}]$. This matrix contains, at the $i$-th row and $j$-th column, the redundancy value between experts $e_i$ and $e_j$, $r_{ij}$. Using this matrix, the MoEITS method selects which experts are least redundant from a collective perspective. For this purpose, an iterative algorithm is defined, which, inspired by the idea of using the interquartile range (IQR) for anomaly detection \cite{Dawson01072011, 10.1007/978-981-10-7563-6_53}, establishes a redundancy limit called $\rho$. This limit is calculated as the mean of the NMI matrix plus its interquartile range multiplied by $\tau$, 
%
%\fxnote{(luisbalru) La definición se inspira en cómo usar el IQR para medir la distancia de un punto al centro de la distribución de la que pertenece.}
%\fixme{Pues ponlo, así se explica la decisión}
%\fxnote{Habitualmente se tiene cota superior e inferior con Q1 - 1.5*IQR, Q3+1.5*IQR y he intentado hacer algo parecido. Lo más purista de acuerdo a esas expresiones sería utilizar la mediana (Q2) pero la distribución de pesos es bastante uniforme en los expertos, por lo que me interesaban las pequeñas alteraciones que se pudieran generar y con la media las capturo.} 
the threshold of the MoEITS method, which modulates the level of simplification applied to the network. The definition of $\rho$ is inspired by the classical use of the IQR to measure the distance of a point from the center of the distribution to which it belongs. Given that the distribution of weights is very uniform across the layers of the experts, the mean is introduced to increase sensitivity to small variations and thus detect subtle but important differences among the experts.
%\fixme{(luisbalru) Incluida explicación anteriormente comentada}

Once the redundancy limit is established, the algorithm iterates over the NMI matrix, $\mathcal{R}$, searching for values that exceed $\rho$. Let us suppose that such a value exists and is located at position $(i,j)$. This implies that experts $e_i$ and $e_j$ exhibit a level of redundancy higher than the required limit, and therefore, one of them could be removed. To decide whether $e_i$ or $e_j$ is the expert to prune, the redundancy they express with the rest is measured. The one that expresses greater redundancy with the other experts will be eliminated. To do this, the mean of the values in the $i$-th and $j$-th rows is calculated (it is also possible by columns, since $\mathcal{R}$ is a symmetric matrix), that is, the level of redundancy of experts $e_i$ and $e_j$ with the remaining experts. 
Let us suppose, without loss of generality, that the most redundant expert of this iteration is $e_i$. Then the $i$-th row and column are removed from the NMI matrix, so that this expert is no longer considered. This process ends when there is no value in the $\mathcal{R}$ that exceeds the redundancy limit. Consequently, considering how $\rho$ is calculated, it can be concluded that high values of $\tau$ generate very reduced simplifications, and conversely, small values ---or values close to 0--- of $\tau$ generate notable reductions in size. More details can be found in Algorithm \ref{alg:expertsimp}.

%\fxnote{He hecho cambios en la redacción de este algoritmo para hacerlo más legible. Aún así no queda muy bien; los comentarios si no caben completos, entonces deberían en la línea siguiente pegados a la derecha. Que no afecten a la vista del algoritmo.}

\begin{algorithm}
	\caption{Building the simplified model}\label{alg:simpmodel}
	\begin{algorithmic}[1]
		\Procedure{Simplified Model}{model, $\tau$}
		\State experts $\gets$ []
		\For{$l$ in model.layers}
		\State NMI\_expert\_block $\gets$ $\mathcal{R}$ for $l$-th expert block \\ \Comment{\textsf{[See subsection \ref*{nmi}]}}
		\State relevant\_experts $\gets$ relevant experts in the $l$-th block \\ \Comment{\textsf{[Algorithm\ref{alg:expertsimp}(NMI\_expert\_block, $\tau$)]}}
		\State experts.add(relevant\_experts)
		\EndFor
		\State simplified model $\gets$ New model including only relevant experts \\\Comment{\textsf{[Relevant experts by layer]}}
        \State simplified model $\gets$ Healing phase (Light retraining) \\\Comment{\textsf{[See subsection \ref*{res}]}}
		\State \textbf{return} simplified model
		\EndProcedure
	\end{algorithmic}
\end{algorithm}

\begin{algorithm}
	\caption{Expert-Block Simplification}\label{alg:expertsimp}
	\begin{algorithmic}[1]
		\Procedure{Expert-Block Simplification}{$\mathcal{R}$, $\tau$} \\ \Comment{\textsf{[Get non-redundant experts within a layer]}}
		\State $m$ $\gets$ mean($\mathcal{R}$)
		\State $iqr$ $\gets$ iqr($\mathcal{R}$) \Comment{\textsf{[Interquartile range]}}
		\State $\rho$ $\gets$ $m+iqr \times \tau$ \Comment{\textsf{[Redundance limit]}}
		\While{($\exists \text{ } r_{ij} \in \mathcal{R} > \rho$)} \\
		\Comment{\textsf{[Determine whether there exists any expert $i$ whose redundancy value with another expert $j$ exceeds $\rho$.]}}
		\State $e_1, e_2 \gets$ the two most similar experts 
		\State Measure the redundancy of $e_1, e_2$ with respect to the remaining experts
		\State Remove the most redundant expert $e_1 \text{ or } e_2$
		\EndWhile
		\State \textbf{return} relevant experts
		\EndProcedure
	\end{algorithmic}
\end{algorithm}

The final step of Algorithm \ref{alg:method} involves assigning the weights of the original model to the simplified one. This represents the distillation of knowledge between the models. To achieve this, in those layers where simplifications have not been carried out, that is, all those unrelated to the experts, there is a direct assignment of weights from the original model to the simplified one. In the case of the experts, it is necessary to adjust the weight matrices by eliminating the rows and columns associated with the experts that have been removed. This applies to both the unit preceding the experts, called the Gate, and the subsequent unit that collects their outputs.

%\fixme{(luisbalru) A ver si os gusta más esta versión} \fxnote{Esta frase no deja claro el ajuste que es necesario hacer. La matriz de la que quitas filas y columnas entiendo que es la NMI matrix. Pero del modelo se quita todo el experto. Entiendo que cada experto se representa por su propia matriz, que se elimina por completo. No hay ajustes de las matrices de expertos.} 

%\fxnote{(luisbalru) No se ajusta, reentrena ni recalibra nada. Los expertos que quedan heredan sus pesos originales y para las unidades que dan información o reciben de los expertos se seleccionan las columnas o filas de las matrices de pesos correspondientes, eliminando aquellas relacionadas con los expertos simplificados}

\subsection{Computational complexity analysis}

This section analyzes the computational complexity of MoEITS. First of all, the following notation is established. Let us denote $L$ as the number of layers in a MoE-LLM model, and $e$ as the number of experts per layer. If the number of experts were not the same,  then let $e= \max\{e_{l_1}, e_{l_2}, \dots, e_{l_L}\}$. Let $k$ be the number of layers per expert. It is again, typically a fixed number. If not, $k$ is defined as the maximum. Finally, let $\nu$ be the maximum number of neurons in the layers of the experts.

As expressed in Algorithm \ref{alg:method}, the MoEITS method comprises three distinct phases. The
initial phase centers on the computation of normalized mutual information metrics
across all experts. This computation is contingent upon the number of layers, the
number of experts (scaling quadratically), the number of Linear layers per expert,
and the number of neurons. Consequently, the
information-theoretic metric extraction phase exhibits a computational efficiency
of $\text{O}(Le^2 \nu k)$.

The second phase involves the construction of a simplified version of the original
model, the details of which are delineated in Algorithms \ref{alg:simpmodel} and \ref{alg:expertsimp}. As can be observed,
Algorithm \ref{alg:simpmodel} exhibits a linear dependency on the number of layers, while Algorithm \ref{alg:expertsimp},
in the worst-case scenario, performs $e^2$ iterations. Consequently, the generation of
a new model also exhibits a linear dependency on $L$ and a quadratic dependency on
$e$. Therefore, the algorithmic complexity of this second phase is $\text{O}(Le^2)$.

Finally, the concluding stage of the algorithm involves assigning and adjusting weights within the simplified model, followed by a retraining phase. This process facilitates the transfer of knowledge from the original, pre-trained model to the simplified model. The computational time efficiency of this phase is $\text{O}(Le^2)$.

In summary, taking into account the partial analysis of each stage of the algorithm, the computational complexity of MoEITS is $\text{O}(Le^2 \nu k)$.

\section{Empirical analysis}
\label{ea}
To assess the effectiveness of MoEITS, a thorough extensive experimentation has been designed and rigorously carried out. It is based on well established benchmarks in the literature. Furthermore, results are compared with those for other state-of-the-art techniques, both in terms of accuracy on each benchmark and the achieved reduction, whether in terms of the number of parameters, sparsity, or quantization.

\subsection{Models, datasets and metrics}

Following the trend in other state-of-the-art proposals and to facilitate comparison with them, the Mixtral $8\times7$B, Qwen1.5-MoE-2.7B and DeepSeek-V2-Lite models are established as a reference for evaluating simplification. Nevertheless, MoEITS is a model-agnostic method, and thus it can be applied to any other MoE.

Similarly, to facilitate task-agnostic evaluation of the method, following the evaluation methodology proposed in \cite{ma2023llmprunerstructuralpruninglarge}, zero-shot task classification is performed across datasets such as BoolQ \cite{clark-etal-2019-boolq}, HelloSwag \cite{zellers-etal-2019-hellaswag}, WinoGrande \cite{10.1145/3474381}, ARC-e and ARC-c \cite{clark2018thinksolvedquestionanswering}. A zero-shot setting implies that the model is evaluated without being explicitly trained on the specific benchmark dataset. This tests the model's general knowledge and reasoning abilities. Table \ref{metrics} contains more details about the tasks, including description, metrics and key focus. Accuracy is used as the prediction metric in all benchmarks. Additionally, the percentage of simplified parameters and sparsity, which refers to the number of weights set to zero after the simplification process, are included as measures to assess the level of simplification.

\begin{table}[htbp]
	\centering
    	\caption{Overview of LLM benchmarks used for evaluating reasoning and comprehension capabilities for simplification methods, showing task descriptions and key focus. For all benchmarks, accuracy is the metric used.}

	\begin{tabular}{l *{3}{c}}
		\toprule
		\textbf{Benchmark} & \textbf{Task Description}                                                                                                                                                                                                                                 & \textbf{Key Focus}                                                                                               \\ \hline
		BoolQ              & \begin{tabular}[c]{@{}c@{}}Question answering where \\ the answer is either ``yes'' or \\ ``no.'' It tests a model's ability \\ to understand a passage\\ and answer a simple boolean \\ question about it.\end{tabular}                                      & \begin{tabular}[l]{@{}c@{}}Reading comprehension, \\ logical reasoning.\end{tabular}                             \\ \hline
		HellaSwag          & \begin{tabular}[c]{@{}c@{}}Commonsense reasoning. \\ Given a context, the model \\ must choose the most likely \\ ending from four options. \\ It's designed to be challenging\\  for AI by including \\ adversarial examples.\end{tabular}               & \begin{tabular}[c]{@{}c@{}}Commonsense inference, \\ context understanding.\end{tabular}                         \\ \hline
		WinoGrande         & \begin{tabular}[c]{@{}c@{}}A large-scale dataset\\  of pronoun resolution \\ problems. The model must \\ determine the referent of \\ a pronoun in a sentence, \\ requiring fine-grained \\ understanding of context \\ and world knowledge.\end{tabular} & \begin{tabular}[c]{@{}c@{}}Pronoun resolution, \\ coreference resolution, \\ commonsense reasoning.\end{tabular} \\ \hline
		ARC-e              & \begin{tabular}[c]{@{}c@{}}A set of multiple-choice\\  science questions \\ designed for elementary \\ school level. It tests \\ a model's ability to \\ understand and reason \\ about scientific concepts.\end{tabular}                                 & \begin{tabular}[c]{@{}c@{}}Scientific reasoning,\\ knowledge application.\end{tabular}                           \\ \hline
		ARC-c              & \begin{tabular}[c]{@{}c@{}}A more difficult subset \\ of the ARC dataset, designed \\ to require deeper reasoning\\ and knowledge than ARC-e.\end{tabular}                                                                                                & \begin{tabular}[c]{@{}c@{}}Advanced scientific \\ reasoning, complex \\ knowledge application.\end{tabular}      \\ \hline
	\end{tabular}
	\label{metrics}
\end{table}

\subsection{Retraining and evaluation setup}
\label{res}
Following the simplification process, a retraining phase is essential to restore performance degraded by pruning. This phase specifically targets expert-related layers, including the gating mechanism and the remaining non-pruned experts, using a training subset of the OpenHermes-2.5 dataset \cite{Balderas_Ruiz2026-tb, huggingfaceTekniumOpenHermes25Datasets}. To ensure computational affordability, a Low-Rank Adaptation (LoRA) strategy \cite{hu2022lora} is implemented with a rank $r=64$, $\alpha=16$, and a dropout rate of 0.1, excluding biases. Notably, this training phase is highly efficient and non-intensive, as it is restricted to a limited number of layers and units, requiring negligible resources compared to full-model training or fine-tuning. Furthermore, these hyperparameters were chosen to provide a generalizable configuration across all experiments and models, rather than being selectively optimized for specific cases.

Training was conducted on an NVIDIA H200 GPU using a batch size of 16, 4 gradient accumulation steps, and a learning rate of $2\times10^{-4}$ under a cosine scheduler. Additional regularization includes a warmup ratio of 0.05, weight decay of 0.05, and the application of NEFTune noise ($\alpha=5.0$) using the AdamW optimizer.

Finally, the evaluation is conducted using the \textit{lm-evaluation-harness} framework \cite{eval-harness}. The assessment employs a 15-shot setting across all benchmarks, with a consistent batch size of 40 for each evaluation cycle.

\subsection{Comparison with state-of-art methods}

The empirical results obtained for the simplification of the MoE-LLMs by MoEITS and the other methods from the state-of-the-art proposals are presented in Table \ref{deepseek-soa-results} (DeepSeek-V2 Lite), Table \ref{qwen-soa-results} (Qwen1.5-2.7B) and Table \ref{mixtral-soa-results} (Mixtral $8\times7$B). This table includes accuracy results for the various benchmarks. Simplification metrics are also included. In particular, the percentage of parameter reduction and sparsity. For both metrics, the higher the value, the greater the simplification. With respect to the state-of-the-art methodologies against which MoEITS is benchmarked, the comparative analysis incorporates the following techniques: MoE-$I^2$ \cite{yang-etal-2024-moe}, MoE-Pruner \cite{xie2024moepruner}, STUN \cite{lee-etal-2025-stun}, DERN \cite{zhou-etal-2025-dropping}, MC-MoE \cite{huang2025mcmoe}, and the method proposed in \cite{lu-etal-2024-experts} by Lu \emph{et al.}~ This selection of comparative methods serves to provide a comprehensive evaluation of MoEITS's performance relative to established and contemporary approaches in the field.

\begin{table}[htbp]
	\centering
    	\caption{Results of pruning DeepSeek-V2 Lite on BoolQ, ARC-e, ARC-c, WinoG (WinoGrande) and HelloS (HellaSwag) benchmarks. Av represents the average of the metrics.  P $\downarrow$ represents the percentage reduction in the number of expert parameters. Best results are highlighted in bold.}

	\begin{tabular}{l *{8}{c}}
		\toprule
		\textbf{Method}        & \textbf{BoolQ} & \textbf{ARC-e} & \textbf{ARC-c} & \textbf{WinoG} & \textbf{HellaS} & \textbf{Av} & \textbf{P $ 	\downarrow$}  \\ \hline
		MoE-$I^2$  \cite{yang-etal-2024-moe}       & 76.79 & 71.8 & 42.58  & 67.64       & 55.16 & 62.79     & 53.98\%           \\ \hline
		MoE-Pruner \cite{xie2024moepruner}   & 76.61 &71.89 & 40.02 & 67.64      & 50.94 & 61.42    & -       \\ \hline
        DERN \cite{zhou-etal-2025-dropping}   & 68.29 &70.9 & \textbf{48.14 }& 53.51        & 44.47 & 57.06     & 22.56\%          \\ \hline
		MoEITS           & \textbf{80.03} &\textbf{77.61} & 43.77 & \textbf{67.72}        & \textbf{70.8} & \textbf{67.99}     & \textbf{55.42\%}        \\ \hline
	\end{tabular}
	\label{deepseek-soa-results}
\end{table}

Firstly, Table \ref{deepseek-soa-results} presents the results for DeepSeek-V2 Lite. The MoEITS method achieves the highest performance across all benchmarks with the exception of ARC-c, where DERN exhibits superior results. Furthermore, regarding model simplification, MoEITS demonstrates significant efficiency, achieving a reduction of over 55\% in expert parameters ($\tau = 1.25$). This represents a decrease of more than 1.5\% compared to MoE-$I^2$ and over 30\% relative to DERN. On average, the proposed method outperforms the best-performing approach by 5\%.

\begin{table}[htbp]
	\centering
    	\caption{Results of pruning Qwen1.5-2.7B on BoolQ, ARC-e, ARC-c, WinoG (WinoGrande) and HelloS (HellaSwag) benchmarks. Av represents the average of the metrics.  P $\downarrow$ represents the percentage reduction  in the number of expert parameters. Best results are highlighted in bold.}

	\begin{tabular}{l *{8}{c}}
		\toprule
		\textbf{Method}        & \textbf{BoolQ} & \textbf{ARC-e} & \textbf{ARC-c} & \textbf{WinoG} & \textbf{HellaS} & \textbf{Av} & \textbf{P $ 	\downarrow$}  \\ \hline
		MoE-$I^2$  \cite{yang-etal-2024-moe}       & 75.08 & 71.68 & 41.13  & 66.54       & 53.08 & 61.5     & \textbf{53.98\% }         \\ \hline
		MoE-Pruner \cite{xie2024moepruner}   & 69.14 & 52.02 & 29.1 & 59.12      & 42.99 & 50.47    & 50\%       \\ \hline
        MoP~\cite{Hu_Zhao_Song_Zhu_Lai_Wang_2026}   & 68.56 & 59.76 & \textbf{44.97} & 52.57      & 56.4 & 56.45   & 33\%       \\ \hline
		MoEITS           & \textbf{75.2} & \textbf{72.3} & 36.01 & \textbf{67.46}   & \textbf{61.27} & \textbf{62.45}     & 52.92\%         \\ \hline
	\end{tabular}
	\label{qwen-soa-results}
\end{table}

Table \ref{qwen-soa-results} illustrates the results for the Qwen1.5-2.7B model. Consistent with previous findings, MoEITS achieves superior performance across the majority of benchmarks, with the notable exception of ARC-c. In this case, MoP attains a score of 44.97\%, compared to 36.01\% for the proposed method. Regarding model simplification, existing state-of-the-art methods yield architectures of comparable size; the margin between MoEITS, with simplification level $\tau=1.25$, and the technique achieving the highest degree of simplification is approximately 1\%. On average, the proposed approach maintains a performance advantage of nearly 1\% over the subsequent leading technique.

\begin{table}[htbp]
	\centering
    	\caption{Results of pruning Mixtral $8\times7$B on BoolQ, ARC-e, ARC-c, WinoG (WinoGrande) and HelloS (HellaSwag) benchmarks. Av represents the average of the metrics.  P $\downarrow$ represents the percentage reduction in the number of expert parameters. S (sparsity) refers to the portion of parameters that are zero. Best results are highlighted in bold.}

	\begin{tabular}{l *{9}{c}}
		\toprule
		\textbf{Method}        & \textbf{BoolQ} & \textbf{ARC-e} & \textbf{ARC-c} & \textbf{WinoG} & \textbf{HellaS} & \textbf{Av} & \textbf{P $ 	\downarrow$} & \textbf{S} \\ \hline
		MoE-$I^2$  \cite{yang-etal-2024-moe}       & 82.6 & 78.2 & 52.2  & 71.5       & 61.1 & 69.1    & 49\%   & -        \\ \hline
		MoE-Pruner \cite{xie2024moepruner}   & 86.0 & 81.9 & 53.3 & 75.5      & 62.3 & 71.8    & 0\%    & 50\%     \\ \hline
		STUN \cite{lee-etal-2025-stun}         & -     & 80.0 & 51.5 & -          & 59.9 & 63.8    & 0\%    & 65\%     \\ \hline
		MC-MoE \cite{huang2025mcmoe}          & 80.6 & 73.1 & 48.4 & 71.3      & 74.9  & 69.6   & -      & -        \\ \hline
		Lu \emph{et al.}~\cite{lu-etal-2024-experts}       & 84.8 & 79.9 & 53.9 & 73.8       & 60.1   & 70.5  & 40\%   & -        \\ \hline
        C-GNN-PRUNE~\cite{Li_Wang_Wang_2026}       & 83.73 & 80.03 & - & 72.37       & 56.16   & 73.1  & \textbf{50\%}   & -        \\ \hline
        MoP~\cite{Hu_Zhao_Song_Zhu_Lai_Wang_2026}       & 86.51 & 82.37 & \textbf{75.43} & 55.80       & 74.28   & 74.88  & 25\%   & -        \\ \hline
		MoEITS           & \textbf{87.33} & \textbf{83.08} & 55.89 & \textbf{80.98}         & \textbf{82.83} & \textbf{78.02}      & 43\%   & -        \\ \hline
	\end{tabular}
	\label{mixtral-soa-results}
\end{table}

Finally, Table \ref{mixtral-soa-results} presents the results for the Mixtral $8\times7$B model. As the primary benchmark for evaluating Mixture-of-Expert simplification techniques, this model entails a more extensive comparison across a broader range of methodologies. In this evaluation, MoEITS achieves superior performance across almost all benchmarks, maintaining a substantial lead in each category and outperforming other methods by an average margin of over 6\% ($\tau = 0.5$). Regarding the degree of simplification, C-GNN-PRUNE produces the most efficient architecture, achieving a 50\% reduction in expert parameters.

 In general terms, MoEITS achieves substantial results in terms of accuracy across the various benchmarks while simultaneously achieving a notable reduction in network size. Therefore, these results enable us to answer research question RQ2, as we can confirm that the presented methodology is effective for the simplification of MoE-LLMs.

\section{Discussion}
\label{discussion}

Considering the results obtained, as detailed in the preceding section, it has been demonstrated that information theory metrics --specifically NMI-- are effective for measuring expert redundancy within an MoE, thereby addressing RQ1. Furthermore, the present study introduces MoEITS, an algorithmic proposal for the simplification of MoE-LLMs (RQ2).

Next, several topics for discussion are presented to further elucidate the key aspects of the proposed method. Specifically, a comparative analysis is introduced, contrasting simplification methods based on pruning, sparsity and quantization. Furthermore, an ablation study is performed, focusing on the parameter $\tau$, for answering RQ3.

\subsection{Sparsity, Quantization and Pruning: Three Paths to Neural Network Efficiency}

The comparison among neural network simplification methods must be conducted according to objective and well-recognized metrics. Firstly, metrics associated with the selected benchmarks are useful for assessing whether the models generated by the simplification methods are capable of solving learning tasks with adequate quality. Then, other fundamental metrics measure the level of simplification, that is, the extent to which the quantity of resources required to solve the referred to tasks has decreased. In this work, techniques based on sparsity as a simplification metric are compared. Other techniques quantify simplification by measuring the reduction in the number of parameters. Sparsity-inducing techniques, such as MoE-Pruner or STUN, operate by imposing constraints or penalties during the training process that encourage a subset of network weights to converge towards zero. This results in a sparse weight matrix, where a significant proportion of connections are effectively nullified. However, it is crucial to recognize that these techniques perform a ``logical simplification'' rather than a physical one. The architectural structure of the network remains unchanged; the number of neurons and connections persists. Instead, the influence of certain connections is rendered negligible through the assignment of zero or near-zero weights. While this process yields a sparse weight matrix, its dimensions, $n \times m$, remain constant. Consequently, during inference, although multiplications involving zero weights do not contribute to the result, the computational overhead associated with these operations and the memory allocation for the entire weight matrix are not eliminated. Thus, sparsity-based methods provide a logical simplification, effectively deactivating certain connections without physically removing them. Unfortunately, they mean no contribution towards the Green AI effort, since the computation resources and energy required are not reduced. 

In contrast, parameter pruning methods, exemplified by techniques such as MoEITS, execute a ``physical simplification.'' These methods involve the explicit removal of redundant or inconsequential neurons and connections from the network architecture. The removal of parameters directly translates to a reduction in the computational and memory requirements. During inference, fewer operations are performed, and less memory is required to store the network's weights. Formally, the transition from the original model to the simplified one represents a structural modification of the network. This modification leads to a decrease in the number of floating-point operations required for inference and a reduction in the memory footprint. 

Finally, while quantization is extraordinary popular and effective for dense models, pruning is a more convenient approach for MoE-LLMs simplification. First of all, the gating network or router in a MoE is highly sensitive. Therefore, quantization can squash the subtle differences in the router's logits. However, pruning solutions drop the redundant experts and, as a result, the routing mechanism remains accurate in its original high-precision format. Besides, pruning methods don't need any custom kernel, dequantization overhead or specific GPU architectures. In contrast, specialized software kernels are required for applying quantization. Finally, LLMs develop outliers, in other words, a small number of weights that are critical for the model's understanding and thinking. Whilst quantization struggles with outliers, pruning techniques avoid this issue \cite{10.5555/3600270.3602468, 10.5555/3618408.3619993, dettmers2023spqrsparsequantizedrepresentationnearlossless}.

\subsection{Ablation study}

\begin{figure}
	\centering
	\includegraphics[width=\textwidth]{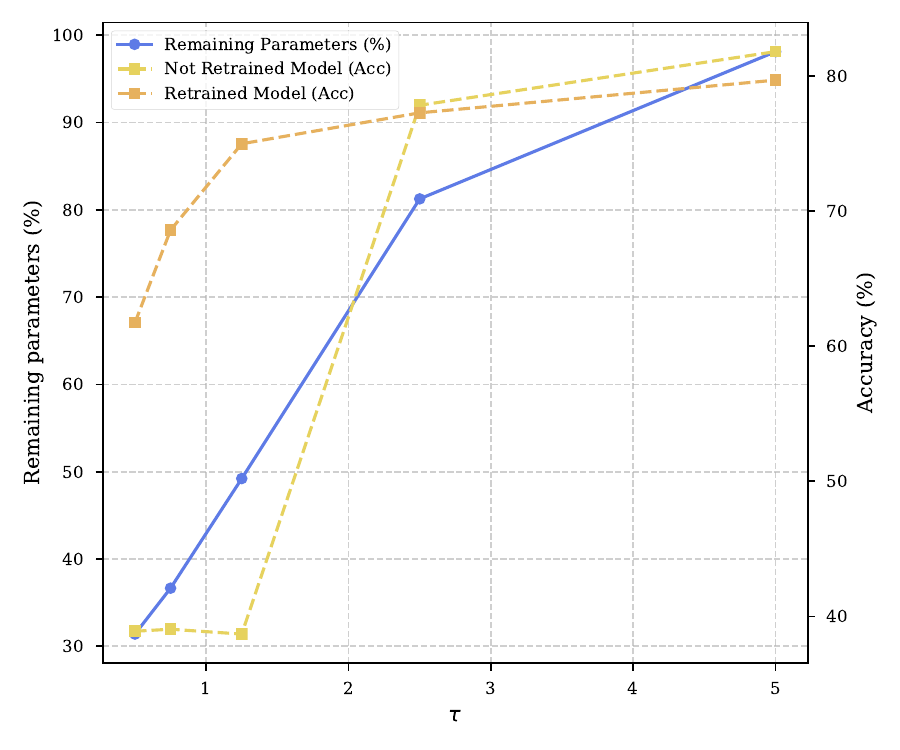}
	\caption{Evolution of the simplified Qwen1.5-2.7B model with MoEITS for different values of $\tau$ ({\color{azulito} number of parameters}, {\color{amarillito} not retrained model accuracy} and {\color{marroncito} retrained model accuracy}). Low $\tau$ values facilitate substantial simplification (up to 70\% parameter reduction for $\tau = 0.5$). The subsequent retraining phase proves critical, restoring accuracy by as much as 20\% relative to the simplification level.}
	\label{fig:as}
\end{figure}

The objective of this section is to analyze the evolution of the generative and reasoning capabilities of Qwen1.5-2.7B when different levels of simplification are applied. To this end, we select, without loss of generality, a benchmark, specifically BoolQ, and apply MoEITS with different threshold values or $\tau$. $\tau$ is used to control the level of simplification applied to the model. A decrease in this parameter corresponds to a more aggressive pruning. Figure \ref{fig:as} shows the evolution of the reasoning capacity of compressed Qwen1.5-2.7B models at different levels of compression. For values of $\tau$ ranging from $0.5$ to $5$, it 
displays in {\color{azulito}blue color} (left axis) the percentage of parameters remaining in the model following the simplification process. The monotonic relationship between $\tau$ and the number of parameters is patent.  

On the other hand, the compressed models' accuracy is also plotted in {\color{amarillito}yellow color} for not retrained simplified versions and in {\color{marroncito} orange color} for the retrained simplified ones (right axis). The accuracy generally follows the upward trend of $\tau$.  As illustrated in the figure, a $\tau$ value of 1.25 achieves substantial simplification, reducing the parameter count by approximately 50\% while maintaining an accuracy of nearly 75\%. Furthermore, despite the increased computational overhead and time requirements, this ablation study demonstrates the necessity of a retraining phase; for non-retrained models reduced to 80\% of their original size ($\tau<2.5$), a drastic loss in reasoning performance is observed.

%\fxnote{Incompleto} \fxnote{(luisbalru) Esta tarde lo hago, aludiendo a RQ3 aquí y en conclusiones}

Regarding research question RQ3, it is evident that for $\tau<1.25$, the model's performance is notably reduced in this case, even though the number of parameters is also significantly reduced (less than 50\% of the original model's parameters). Consequently, parametric pruning algorithms, such as MoEITS, prove particularly valuable in identifying the optimal trade-off between performance and model size. Experiments like this show that a model can be readily generated to meet the hardware constraints of the target deployment device, while leveraging all available information regarding accuracy.

%\fixme{(luisbalru) añadido ablation study}
\section{Conclusions}
\label{conclusions}

Large Language Models (LLMs), and particularly Mixture-of-Experts LLMs (MoE-LLMs), are models that have revolutionized both industry and academia by providing cutting-edge and effective solutions across virtually all disciplines of Machine Learning. However, not only has their training required vast amounts of data and computational capacity, but their deployment is also so infrastructure-demanding that it is within the reach of only a few companies and research centers. In this paper, MoEITS, a simplification method for MoE-LLMs based on information theory, is proposed. Specifically, it employs normalized mutual information (NMI) to detect redundancy among the experts, enabling the removal of the most redundant ones and, consequently, significantly reducing the network with minimal loss in predictive quality.

To assess the effectiveness of MoEITS, a comprehensive analysis was conducted. On the one hand, from a theoretical perspective, it was established that its computational complexity is $\text{O}(Le^2\nu^2k)$, where $L$ represents the number of blocks, $e$ indicates the number of experts per block, $k$ represents the number of  layers per expert and $\nu$ the maximum number of neurons in the layers of the experts. On the other hand, a thorough experimentation was designed and carried out. It includes the most established benchmarks in the literature, both at the model level using Mixtral $8\times7$B, DeepSeek-V2 Lite and Qwen1.5-2.7B and at the task learning level. The results demonstrate that MoEITS is more effective and efficient than other state-of-the-art methods. Consequently, it can be asserted that NMI is a useful metric for measuring redundancy in experts (RQ1) and that MoEITS is a versatile tool for the simplification of MoE-LLMs (RQ2). Finally, an ablation study was conducted on the $\tau$ parameter to address (RQ3). The results demonstrate that beyond a certain simplification threshold, the model's reasoning capacity undergoes a precipitous decline. Therefore, conducting such studies is crucial for determining the threshold value that aligns with the infrastructure requirements of the intended deployment environment.

\section*{Acknowledgment}

This research has been partially supported by Proyecto PID2023-151336OB-I00 financiado por MICIU/AEI /10.13039/501100011033 y por FEDER, UE.

\section*{Data Availability}

No proprietary data is required to run this code. The healing (fine-tuning) phase of MoEITS utilizes a specific subset (training) of the publicly available OpenHermes dataset to recover model performance after pruning. 

To ensure strict reproducibility and permanent archiving, the exact data subset used in our experiments has been deposited in Zenodo.

\begin{itemize}
    \item Zenodo Archive \cite{Balderas_Ruiz2026-tb}
    \item Original Dataset: OpenHermes 2.5 \cite{huggingfaceTekniumOpenHermes25Datasets}
\end{itemize}

%% If you have bib database file and want bibtex to generate the
%% bibitems, please use
%%
%%  \bibliographystyle{elsarticle-num} 
%%  \bibliography{<your bibdatabase>}

%% else use the following coding to input the bibitems directly in the
%% TeX file.

%% Refer following link for more details about bibliography and citations.
%% https://en.wikibooks.org/wiki/LaTeX/Bibliography_Management

\bibliographystyle{plain} 
\bibliography{citas}
\end{document}